\documentclass{article}

\usepackage[utf8]{inputenc} % allow utf-8 input
\usepackage[T1]{fontenc}    % use 8-bit T1 fonts
\usepackage{hyperref}       % hyperlinks
\usepackage{url}            % simple URL typesetting
\usepackage{booktabs}       % professional-quality tables
\usepackage{amsfonts}       % blackboard math symbols
\usepackage{nicefrac}       % compact symbols for 1/2, etc.
\usepackage{microtype}      % microtypography

\usepackage{lipsum}

\usepackage{wrapfig}

\usepackage{graphicx}

\usepackage{subcaption}
\usepackage{mwe}

\usepackage{xspace}

\usepackage{latexsym}
\usepackage{amsmath}
\usepackage{amssymb}

\usepackage{comment}
\usepackage{enumitem}

\usepackage[noend]{algpseudocode}
\usepackage{algorithm}

\newcommand{\TA}[1]{\textcolor{magenta}{ #1}}
\newcommand{\TAEN}[1]{\endnote{\textcolor{magenta}{ #1}}}
\newcommand{\TAFN}[1]{\footnote{\textcolor{magenta}{ #1}}}
\newcommand{\MD}[1]{\textcolor{cyan}{ #1}}
\newcommand{\MDEN}[1]{}
\newcommand{\MDFN}[1]{}
\usepackage{outlines}
\usepackage{enumitem}

\usepackage{xcolor}

\DeclareMathOperator{\E}{\mathbb{E}}

\newcommand{\pit}{\pi_\theta}
\newcommand{\Pl}{P_\lambda}
\newcommand{\pl}{p_\lambda}
\newcommand{\CE}{\text{CE}}
\newcommand{\nabt}{\nabla_\theta}
\newcommand{\ptrue}{p_{true}}

\newcommand{\dpgon}{DPG$_\text{on}$}
\newcommand{\dpgoff}{DPG$_\text{off}$}

\newcommand{\SupMat}{Sup.~Mat.}

\newcommand{\rs}{\textit{rs}\xspace}
\newcommand{\snis}{\textit{snis}\xspace}

\usepackage{endnotes}
\makeatletter
\renewcommand*\makeenmark{\hbox{\textsuperscript{A\@arabic{\theenmark}}}}
\makeatother

\usepackage[export]{adjustbox}

\usepackage{tikz}
\usetikzlibrary{positioning}

\DeclareMathOperator*{\argmax}{argmax}
\DeclareMathOperator*{\argmin}{argmin}

\usepackage[showseconds=false,showisoZ=false]{datetime2}

\title{Distributional Reinforcement Learning for Energy-Based Sequential Models}
\author{\textit{Tetiana Parshakova \qquad Jean-Marc Andreoli \qquad Marc Dymetman}\\[2mm]
	{{\hspace{-2.2cm} Stanford University\thanks{\ Work done while at Naver Labs Europe.} \hspace{2.7cm} Naver Labs Europe}}\\[2mm]
	{\small{\tt \hspace{0.1cm} tetianap@stanford.edu \hspace{0.6cm} \{jean-marc.andreoli,marc.dymetman\}@naverlabs.com}}
}

\date{}

\begin{document}

\maketitle

\begin{abstract} Global Autoregressive Models (GAMs) are a recent proposal \cite{gams-conll-2019} for exploiting global properties of sequences for data-efficient learning of seq2seq models. In the first phase of training, an Energy-Based model (EBM) \cite{lecun_tutorial_2006} over sequences is derived. This EBM has high representational power, but is unnormalized and cannot be directly exploited for sampling. To address this issue \cite{gams-conll-2019} proposes a distillation technique, which can only be applied under limited conditions. By relating this problem to Policy Gradient techniques in RL, but in a \emph{distributional} rather than \emph{optimization} perspective, we propose a general approach 
applicable to any sequential EBM. Its effectiveness is illustrated on GAM-based experiments.
\end{abstract}

%%%%%% INTRODUCTION %%%%%%
% \input{introduction.tex}

\section{Introduction}

The mainstream autoregressive sequence models 
\cite{hochreiter1997long,SutskeverVL14,gehring2017convolutional,Vaswani}) form a subclass of sequential energy-based  models (sequential EBMs) \cite{lecun_tutorial_2006}. While the former are locally normalized and easy to train and sample from, the latter allow global constraints, greater expressivity, and potentially better sample efficiency, but lead to unnormalized distributions and are more difficult to use for inference and evaluation. We exploit a recently introducaked class of energy-based models, Global Autoregressive Models (GAMs) \cite{gams-conll-2019}, which combine a locally normalized component (that is, a first, standard, autoregressive model, denoted $r$) with a global component and use these to explore some core research questions about sequential EBMs, focussing our experiments on synthetic data for which we can directly control experimental conditions.
We dissociate the (relatively easy) task of learning from the available data an energy-based \emph{representation} (Training-1), from the more challenging task of \emph{exploiting} that representation to produce samples or evaluations (Training-2).

In this paper, we provide a short self-contained introduction to GAMs and to their two-stage training procedure. However our main focus is about Training-2. For that task \cite{gams-conll-2019} proposed a Distillation technique to project the Energy-Based representation (denoted by $\Pl$) obtained at the end of \mbox{Training-1} into a final autoregressive model (denoted $\pit$), with better test perplexity than the initial $r$, but this technique was limited to cases where it was possible to sample from $\Pl$ at training time. 
One key observation of the current submission is that Training-2, considered as the general problem of deriving an autoregressive model from an energy-based model (not necessarily obtained through Training-1) has strong similarities with the training of policies in Reinforcement Learning (RL), but in a \emph{distributional}  rather than in an \emph{optimization} perspective as in standard RL. We then propose a distributional variant of the Policy Gradient technique (Distributional Policy Gradient: DPG) which has wider applicability than distillation. We conduct GAM-based experiments to compare this technique with distillation, in synthetic data conditions where distillation is feasible, and show that DPG works as well as distillation.
In both cases, in small data conditions, the policies (aka autoregressive) models $\pit$ obtained at the end of the process are very similar and show strong perplexity reduction over the standard autoregressive models.

Section 2 provides an overview of GAMs. Section 3 explains the training procedure, with focus on EBMs and relations to RL. Section 4 presents experiments and results. For space reasons we use the Supplementary Material (\SupMat) to provide some details and to discuss related work.

% \MD{Add some outline here, also saying something about \SupMat{} in particular that it provides some related work. Could also copy AAAI intro to \SupMat{} ???}

\section{Model}
\label{sec:model}

\subsection{Background}
\paragraph{Autoregressive models (AMs)} These are currently the standard for neural seq2seq processing, with such representatives as RNN/LSTMs \cite{hochreiter1997long,SutskeverVL14}, ConvS2S \cite{gehring2017convolutional}, Transformer \cite{Vaswani}). Formally, they are defined though a distribution $r_\eta(x|C)$, where $x$ is a target sequence to be generated, and $C$ is a context, with $r_\eta(x|C) \doteq \prod_i s_\eta(x_i|x_1,\ldots,x_{i-1}, C)$, and
where each $s_\eta(x_i|x_1,\ldots,x_{i-1}, C)$ is a normalized conditional probability over the next symbol of the sequence, computed by a neural network (NN) with parameters $\eta$. The local normalization of the incremental probabilities implies the overall normalization of the distribution $r_\eta(x|C)$. In RL terminology, AMs can also be seen as \emph{policies} where actions are symbols and states are sequence prefixes. 

\paragraph{Energy-Based Models (EBMs)} EBMs are a generic class of models, characterized by an energy function $U_\eta(x|C)$ computed by a neural network parametrized by $\eta$ \cite{lecun_tutorial_2006}. Equivalently, they can be seen as directly defining a potential (an unnormalized probability distribution) $P_\eta(x|C) = e^{-U_\eta(x|C)}$, and indirectly the normalized distribution $p_\eta(x|C) = 1/Z_{\eta}(C)\ P_\eta(x|C)$, with $Z_{\eta}(C) = \sum_x P_\eta(x|C)$. 
%  
\begin{comment}
    A fundamental property of these models is that, for max-likelihood training, the SGD updates can be computed through the formula:\footnote{See \citep[p. 15]{lecun_tutorial_2006}, and [SM] for a derivation.} 
    \begin{align} \label{eq:EBM-SGD}
    \nabla_\eta \log p_\eta(x|C) &= \nabla_\eta \log P_\eta(x|C)\\
    &- E_{x \sim p_\eta(\cdot|C)} \nabla_\eta \log P_\eta(x|C), \nonumber
    \end{align}
    which, in principle, reduces the problem of \emph{training} with unnormalized potentials to the problem of \emph{sampling} from them.
\end{comment}
% 
Here we will identify an EBM with its potential (the $P_\eta$ form) and be  concerned exclusively with sequential EBMs, that is, the case where $x$ is a sequence.%
%\MDEN{Generality of SEBMs: many general objects can be decomposed in terms of a \emph{sequence of decisions}. Then providing an EBM over the objects can be seen as providing an SEBM over sequences! This means that the approach taken in this paper may have wider applications than to text processing.}

\subsection{GAMs} 
We employ a specific class of sequential EBMs, \emph{Global Autoregressive Models} (GAMs), which we summarize here (for details please see \cite{gams-conll-2019}).
GAMs exploit both local autoregressive properties as well as global properties of the sequence $x$. A GAM is an unnormalized potential $P_\eta(x|C)$ over $x$, parametrized by a vector $\eta = \eta_1 \oplus \eta_2$, which is the product of two factors:
\begin{equation}\label{eq:GAM}
P_\eta(x|C) = r_{\eta_1}(x|C) \cdot e^{\langle\lambda_{\eta_2}
(C),\ \phi(x;C)\rangle}.   
\end{equation}
Here the factor $r_{\eta_1}(x|C)$ is an autoregressive model for generating $x$ in the context $C$, parametrized by $\eta_1$. The factor 
$e^{\langle\lambda_{\eta_2}(C),\ \phi(x;C)\rangle}$
on the other hand, is a \emph{log-linear} potential \cite{Jebara2013}, where $\phi(x;C)$ is a vector of predefined real features of the pair $(x,C)$, which is combined by a scalar product with a real vector $\lambda_{\eta_2}(C)$ of the same dimension, computed by a network parametrized by $\eta_2$. The normalized distribution associated with the GAM is $p_\eta(x|C) = \frac{P_\eta(x|C)}{Z_\eta(C)}$, where $Z_\eta(C) = \sum_x P_\eta(x|C)$.

The motivations for GAMs are as follows. The first factor guarantees that the GAM will have at least the same effectiveness as standard autoregressive models to model the local, incremental, aspects of sequential data.%
%\MDEN{Should we explain this more ?}
The second factor can be seen as providing a ``modulation'' on the first one. While we could have chosen any energy-based potential for that factor, the log-linear form has several advantages. First, the features $\phi(x;C)$ provide prior knowledge to the model by drawing its attention to potentially useful global sequence properties that may be difficult for the AM component to discover on its own. Second, log-linear models enjoy the following important property: at maximum likelihood, the features expectations according to the model and to the data are equal (``moment matching'' property). 

In our experiments, we focus on a simple unconditional (language modelling) version of GAMs, of the form:
    \begin{align}
    P_\lambda(x) &\doteq r(x) \cdot e^{\langle \lambda,\ \phi(x)\rangle}, \label{eq:exp-family}
    \end{align}
where the autoregressive factor $r = r_{\eta_1}$ is first learnt on the training dataset of sequences $D$ and then kept fixed, and where the parameter vector $\lambda$ is then trained on top of $r$, also on $D$. We denote by $p_\lambda(x)$ the normalized distribution associated with $P_\lambda(x)$.%
%\MDEN{Perhaps clarify why $\lambda$ and not $\lambda_{\eta_2}$.}
%\MDEN{Perhaps mention somewhere that we could train $r_{\eta_1}$ jointly with $\lambda$.}

\section{Training}
\label{sec:training}

\begin{wrapfigure}{R}{0.3\textwidth}
\vspace{-5mm}
\begin{minipage}{0.3\textwidth}
\centering
\begin{figure}[H]
    \scalebox{0.9}{
    \begin{tikzpicture}
        \node (r) at (0,0.5) {$r(x)$};
        \node (pi) at (0,-0.5) {$\pi_\theta(x)$};
        \node (P) at (3,0) {$P_\lambda(x)$};
        \draw[->,>=latex] (r) to[bend left=45] node[below of=mid, pos=0.4, yshift=4mm] {Training-1} (P);
    \draw[->,>=latex] (P) to[bend left=45] node[above of=mid, pos=0.6, yshift=-4mm] {Training-2} (pi);
    \end{tikzpicture}
    }
    \caption{\label{f:tstage} Two-stage training. At the end of the process, we compare the perplexities of $r$ and $\pit$ on test data: $CE(T,r)$ vs. $CE(T,\pit)$.}
\end{figure}
\end{minipage}
\vspace{2mm}
\end{wrapfigure}
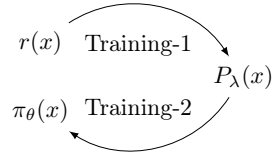

We assume that we are given a training data set $D$ (resp. a validation set $V$,  a test set $T$) of sequences $x$, and a finite collection of real-valued feature functions $\phi_1,\ldots,\phi_k$. 
The GAM training procedure then is performed in two stages (see Fig.~\ref{f:tstage}).

\subsection{Training-1: from data to energy-based representation} This phase consists in training $P_\lambda$ by max-likelihood (ML) on $D$. We start by training an AM $r = r_{\eta_1}$ (our initial policy) on $D$, in the standard way. We then fit the log-linear weight vector $\lambda$ to the data. In order to do that, we denote by $\log p_{\lambda}(D)$ the log-likelihood of the data, and perform SGD over $\lambda$ by observing that (\ref{eq:exp-family}) implies:
\begin{small}
\begin{equation} 
\nabla_\lambda\log p_{\lambda}(D) = |D| \cdot [E_{x \sim p_D(x)}\ \phi(x) - E_{x \sim p_\lambda(\cdot)}\ \phi(x)], 
\end{equation}
\end{small}
\hspace{-0.3em}%
where $E_{x \sim p_D(x)} \ \phi(x)$ (resp. $E_{x \sim p_\lambda(\cdot)}\ \phi(x)$) denotes the expectation (aka moment) of the feature vector relative to the data (resp. to the model).  
The first moment can be directly computed from the data, but the second moment requires more effort. The most direct way for estimating $E_{x \sim p_\lambda(\cdot)}\ \phi(x)$ would be to produce a random sample from $p_\lambda(\cdot)$ and to compute the mean of $\phi(x)$ over this sample. In general, when starting from an unnormalized $\Pl$ as here, obtaining samples from $\pl$ can be difficult. One approach consists in applying a Monte-Carlo sampling technique, such as Rejection Sampling (\rs) \cite{Robert:2005:MCS:1051451}, and this is one of two techniques that can be applied in the experimental conditions both of \cite{gams-conll-2019} and of this paper. However rejection sampling is feasible only in situations where reasonable upper-bounds of the ratio $P(x)/q(x)$ (for $q$ a proposal distribution) can be derived.\footnote{More sophisticated MCMC sampling techniques with broader applicability exist \cite{Robert:2005:MCS:1051451}, but they are typically difficult to control and slow to converge.} This is why \cite{gams-conll-2019} proposes another technique of wider applicability, Self-Normalized Importance Sampling (\snis) \cite{owen_adaptive_2017,y._bengio_adaptive_2008}.This technique directly estimates the expectation $E_{x \sim p_\lambda(\cdot)}\ \phi(x)$ without requiring samples from $\pl$.
%it benefits from the fact that $r$ is a reasonable approximation to the true distribution, and can serve as an efficient IS proposa.

\subsection{Training-2: from energy-based representation to distributional policy} 
The output of the previous stage is an unnormalized EBM, which allows us to compute the potential $P(x) = P_\lambda(x)$ of any given $x$, but not directly to compute the partition function $Z = \sum_x P(x)$ nor the normalized distribution $p(x) = 1/Z\ P(x) = p_\lambda(x)$ or to sample from it.%
\footnote{In our discussion of Training-2, to stress the generality of the techniques employed, we will use $P(x)$ to denote \emph{any} EBM potential over sequences, and $p(x) = 1/Z\ P(x)$, with $Z = \sum_x P(x)$, to denote the associated normalized distribution. Whether $P(x)$ is obtained or not through Training-1 in a GAM-style approach is irrelevant to this discussion.}
In RL terms, the score $P(x)$ can be seen as a \emph{reward}. The standard RL-as-optimization view would lead us to search for a way to maximize the expectation of this reward, in other words for a policy $\pi_{\theta^{*}}$ with $\theta^{*} = \argmax_\theta \E_{x\sim \pit(\cdot)} P(x)$, which would tend to concentrate all its mass on a few sequences.%
%\footnote{While this view could make sense in some seq2seq applications, it does not for language modelling, as explained in the introduction.%\MDEN{Actually this view is also interesting. $\Pl$ is a way to obtain the reward in a ML way, \emph{even if we then used this reward in the standard RL sense!}}}%
%\MDEN{Namely the set\footnote{Often reduced to a singleton.} $\argmax_x P(x)$. We could note that the ability for the policy to be concentrated on this set is conditioned on the capacity of the $\pit$ model family.}%

By contrast, our RL-as-sampling (distributional) view consists in trying to find a policy $\pi_{\theta^{*}}$ that approximates the distribution
 $p$ as closely as possible, in terms of cross-entropy $CE$. We are thus trying to solve $\theta^{*} = \argmin_\theta \CE(p,\pit)$, with $CE(p,\pit) = - \sum_x p(x) \log \pit(x)$. We have:
\begin{small}
	\begin{align}
	\nabt\ \CE(p,\pit) &= - \sum_x p(x)\ \nabt \log \pit(x) 
	= - \E_{x \sim p(\cdot)} \nabt \log \pit(x). \label{eq:sgd-1}
	\end{align}
\end{small}
We can apply \eqref{eq:sgd-1} for SGD optimization, using different approaches.

The simplest approach, \textbf{Distillation}, can be employed in situations where we are able to draw, in reasonable training time, a large number of samples $x_1, \ldots, x_K$ from $p$. We can then exploit \eqref{eq:sgd-1} directly to update $\theta$, which is in fact equivalent to performing a standard supervised log-likelihood SGD training on the set $\{x_1, \ldots, x_K\}$. This is the approach to Training-2 taken in \cite{gams-conll-2019}, using rejection sampling at training time for obtaining the samples, and then training $\theta$ on these samples to obtain a final AM $\pit$ which can be used for efficient sampling at test time and for evaluation. The advantage of this approach is that supervised training of this sort is very succesful for standard autoregressive models, with good stability and convergence properties, and an efficient use of the training data through epoch iteration.\footnote{Epoch iteration might actually be seen as a form of ``experience replay'', to borrow RL terminology \cite{mnih2015human}.} 
However, the big disadvantage is its limited applicability, due to restrictive conditions for rejection sampling, as explained earlier.

A central contribution of the present paper is to propose another class of approaches, which does not involve sampling from $p$, and which relates to standard techniques in RL. We can rewrite the last formula of \eqref{eq:sgd-1} as:
\begin{small}
	\begin{align}
	\sum_x p(x)\ \nabt \log \pit(x) = \frac{1}{Z} \E_{x \sim \pit(\cdot)} \frac{P(x)}{\pit(x)} \nabt \log \pit(x). \label{eq:dpgon}
	\end{align}
\end{small}
This formula is very close to the vanilla formulation (aka REINFORCE \cite{Williams92Reinforce}), we have a reward $R(x)$ and we try to maximize the expectation $\E_{x \sim \pit(\cdot)} R(x)$. It can be shown \cite{Sutton2018} that $\nabt \E_{x\sim \pit(\cdot)} R(x) = \E_{x\sim \pit(\cdot)} R(x)\ \nabt \log \pit(x)$. Thus, in the RL case, an SGD step consists in sampling $x$ from $\pit$ and computing $R(x) \nabt \log \pit(x)$, while the SGD step in \eqref{eq:dpgon} only differs by replacing $R(x)$ by $\frac{P(x)}{\pit(x)}$.%
\footnote{The constant factor $1/Z$ can be ignored here: during SGD, it has the effect of rescaling the learning rate.}
We will refer to the approach \eqref{eq:dpgon} through the name \textbf{Distributional Policy Gradient} (on-policy version) or \dpgon{} (``on-policy'' because the sampling is done according to the same policy $\pit$ that is being learnt).

An off-policy variant \dpgoff{} of \eqref{eq:dpgon} is also possible. Here we assume that we are given some fixed proposal distribution $q$ and we write: 
\begin{small}
	\begin{align}
	\sum_x p(x)\ \nabt \log \pit(x) = \frac{1}{Z} \E_{x \sim q(\cdot)} \frac{P(x)}{q(x)} \nabt \log \pit(x). \label{eq:dpgoff}
	\end{align}
\end{small}
Here the sampling policy $q$ is different from the policy being learnt, and the formula \eqref{eq:dpgoff} represents a form of Importance Sampling, with $q$ the proposal, typically chosen to be an approximation to $p$.

We did some initial experiments with \dpgon{}, but found that the method had difficulty converging, probably due in part to the instability induced by the constant change of sampling distribution (namely $\pit$). A similar phenomenon is well documented in the case of the vanilla Policy Gradient in standard RL, and techniques such as TRPO \cite{schulman2015trust} or PPO \cite{schulman2017proximal} have been developed to control the rate of change of the sampling distribution. In order to avoid such instability, we decided to focus on \dpgoff{}, based on Algorithm~\ref{al:DPG} below.  

\begin{wrapfigure}{r}{0.5\textwidth}
\vspace{-8mm}
\begin{minipage}{0.5\textwidth}
\begin{algorithm}[H]
\caption{\ \dpgoff} \label{al:DPG}
\begin{small}
\begin{algorithmic}[1]
\Require $P$, initial policy $q$ % \Comment{in GAMs: $P=\Pl$ and $\pi_{\theta_0} = r$}
\State $\pit \gets q$
\For{each iteration}
\For{each episode}
    \State sample $x$ from $q(\cdot)$
    \State $\theta \gets \theta + \alpha^{(\theta)} \frac{P(x)}{q(x)}\ \nabla_\theta \log \pi_\theta(x)$ % \Comment{$\alpha^{(\theta)}$: learning rate}
\EndFor
\If{$\pi_\theta$ is superior to $q$} % \Comment{in terms of validation perplexity}%, see text}
    \State $q \gets \pi_\theta$
\EndIf
\EndFor
\Ensure $\pi_\theta$
\end{algorithmic}
\end{small}
\end{algorithm}
\end{minipage}
\vspace{-5mm}
\end{wrapfigure}

In this algorithm, we suppose that we have as input a potential function $P$, and an initial proposal distribution $q$; in the case of GAMs, we take $P=\Pl$ and a good $\pi_{\theta_0}$ is provided by $r$. We then iterate the collection of episodes $x$ sampled with the \emph{same} $q$ (line 4), and perform SGD updates (line 5) according to \eqref{eq:dpgoff} ($\alpha^{(\theta)}$ is the learning rate). We do update the proposal $q$ at certain times (line 7), but only based on the condition that the current $\pit$ is superior to $q$ in terms of perplexity measured on the validation set $V$, thus ensuring a certain stability of the proposal.

This algorithm worked much better than the \dpgon{} version, and we retained it as our implementation of DPG in all our experiments.

\section{Experiments}

%\MD{\textbf{TODO (TP):} some material similar to lines 398-499 of CONLL}
%\MD{\textbf{TODO (TP):} Somewhere, perhaps not exactly her https://www.overleaf.com/4862588695tpgxhtmsfvrte: some material similar to lines 534-543 of CONLL}

In order to assess the validity of our approach, we perform experiments under controllable conditions based on synthetic binary sequences. Our setup is similar to that of \cite{gams-conll-2019}. We generate datasets $D,V,T$ of binary sequences according to a underlying process $\ptrue$. This process produces random ``white noise'' binary strings with fixed length $n=30$ that are filtered according  to whether they contain a specific, fixed, substring (``motif") anywhere inside the sequence. The interest of such a process is that one can efficiently generate datasets (by implementing the filtering process through a probabilistic finite-state automaton) and also directly compute the theoretical entropy (perplexity) of the process (see \cite{gams-conll-2019}). Also, \cite{gams-conll-2019} observed that $\ptrue(x)$ could be well approximated by a standard autoregressive model $r(x)$ when the training dataset was large.

In these experiments, we employed a GAM architecture according to \eqref{eq:exp-family}, using a fixed set of five binary features\footnote{We also did experiments involving two continuous features ($M$ and $v$) assessing length, see \ref{supmat:length-features} in \SupMat}: one feature corresponding to the presence/absence of the motif in the candidate sequence, and four ``distractor'' features with no (or little) predictive value for the validity of the candidate sequence (this feature set, using \cite{gams-conll-2019} notation, is denoted in the figures by the mask $ft=1001111$). We vary the motifs $m$ used, the size of the training set $D$, and the seeds employed.

Our implementation is based on PyTorch \cite{paszke2017automatic}, with policies (i.e. autoregressive models $r$ and $\pit$) implemented as LSTM models over the vocabulary  $\{0, 1, \langle \textrm{EOS} \rangle \}$, with each token represented as a one-hot vector.%
%
%\footnote{Implementation available at \texttt{\scriptsize https://github.com/parshakova/GAMS-Distributional-RL}}

% (from $500$ to $2\cdot10^4$), while the size of the test set $T$ is fixed at $5\cdot 10^3$.\MD{Size of $V$ ?}

The specific experimental setup that we use, due to the nature of the features (binary features or length features $M,v$), permits to perform Training-2 through distillation (the method used in \cite{gams-conll-2019}). In these experiments, we want to confirm that the more generally applicable DPG method works equally well. We do so by varying the training dataset size $D$ and by computing the test perplexity (cross-entropy) of the $\pi_\theta$ obtained at the end of Training-1 + Training-2, and then checking that both distillation and DPG lower this perplexity relative to that of the initial $r$, under small data conditions (data efficiency). But we also confirm that in Training-2, both distillation and DPG are able to almost perfectly approximate the EBM $\Pl$ obtained at the end of Training-1 (that is, to approximate the associated normalized $\pl$); in other words, when $\Pl$ is able to model the $\ptrue$ accurately (which depends on both the quality of the initial $r$ and on the ability of the features to fit the underlying process), then DPG is able to produce a $\pit$ that accurately represents $\ptrue$.

\paragraph{Overall Training: Distillation vs. DPG}

\begin{wrapfigure}{R}{0.5\linewidth}
    \vspace{-10mm}
    \begin{center}
    \includegraphics[ width=1.0\linewidth]{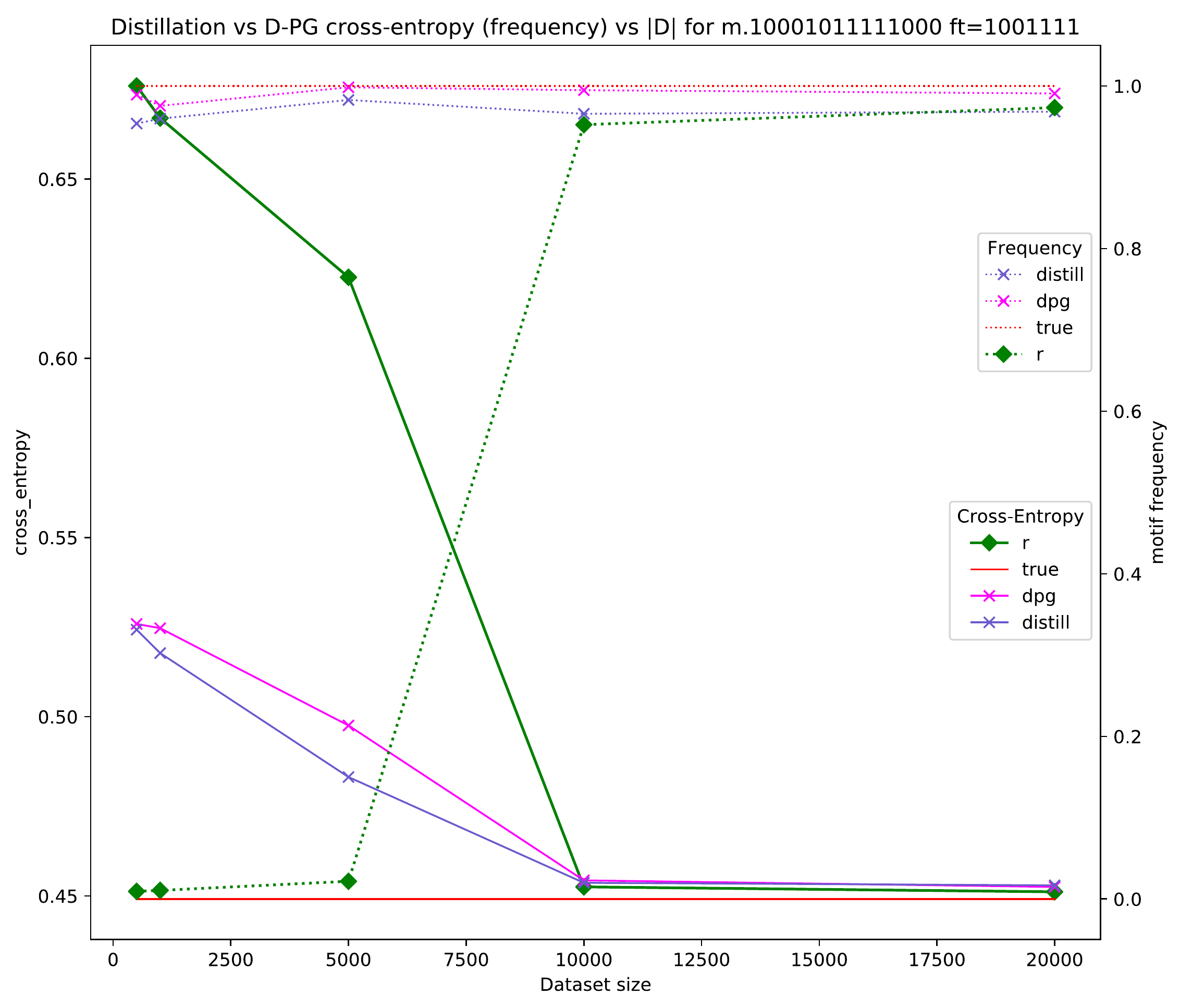} 
    \caption{\label{f:dpg_distill_m2s1} Distillation vs. DPG}
    \end{center}
    \vspace{-7mm}
\end{wrapfigure}

We consider a situation where Training-1 is done through \snis, but Training-2 is done either through Distillation or through DPG (i.e. \dpgoff).
\figurename~\ref{f:dpg_distill_m2s1} illustrates this case. Here the motif, feature vector, and seed are fixed, but the training size $|D|$ varies from $500$ to $2\cdot10^4$) (the size of the test set $T$ is fixed at $5\cdot 10^3$).

The solid lines represent the cross entropies of the final $\pit$ relative to the test set, with the scale located on the left side of the figure, while the dashed lines are the frequencies of the motif  $m$ (computed on $2000$ strings sampled from $\pit$) with the corresponding scale on the right. We distinguish two versions of Training-2, one based on distillation (\texttt{distill}), the other on DPG (\texttt{dpg}). 

First consider the points above $|D|=5000$, and the solid lines: for both \texttt{distill} and \texttt{dpg}, we have $CE(T,r) \gg CE(T,\pi_\theta)\approx H(\ptrue)$: $\pit$ is more data efficient than the initial AM $r$. For smaller data conditions, the tendency is even stronger, while larger $D$ lead to an initial $r$ which is already very good, and on which the two-stage training cannot improve.

Similar conclusions hold for the motif frequencies of $\pit$ compared to $r$: in small data conditions, the motif is much more frequently present when using $\pit$.

Finally, comparing \texttt{distill} and \texttt{dpg}, we see that the performances are very comparable, in this case with a slight advantage of \texttt{distill} over \texttt{dpg} in perplexities but the reverse in motif frequencies.

\paragraph{Effectiveness of DPG in approximating $p$}

To emphasize the performance of DPG in Training-2 (that is, its effectiveness at finding a distributional policy $\pit$ for an EBM representation $P(x)$), independently of the quality of Training-1), we considered two alternatives for $P$. The first one took $P=\Pl$, the energy-based model obtained from Training-1. In our specific experimental conditions, we were able to accurately estimate (via importance sampling) the partition function $Z$ and therefore to compute the cross entropy $CE(T, p_\lambda)$, and to compare it with $CE(T, \pit)$: they were extremely close. We confirmed that finding by considering an alternative where $P$ was defined \emph{a priori} in such a way that we could compute $p$ and $CE(T, p)$ exactly, observing the same behavior. Details are provided in \SupMat{} \ref{supmat:dpg-effectiveness}.

%\paragraph{Results}

\begin{table*}[t!]
\footnotesize
\begin{center}
\footnotesize
\caption{\label{t:stats_dpg_distill_val} Statistics over: $motif \in \{1000101000101,1011100111001$, $10001011111000\}, ft \in \{ 1001111,Mv1001111\}$, $seed \in \{1234, 4444\}$.}
\begin{tabular}{lcccccccc} \footnotesize
%\hline
\rule{0pt}{3ex}$|D|$ & $\frac{\text{CE}(T, \pi_\theta^{dpg})}{\text{CE}(T, \pi_\theta^{dis})}$ & $\frac{\text{mtf\_frq}(\pi_\theta^{dpg})}{\text{mtf\_frq}(\pi_\theta^{dis})}$ & $\frac{\text{CE}(T, \pi_\theta^{dpg})}{\text{CE}(T, r)}$ & $\frac{\text{CE}(T, \pi_\theta^{dpg})}{H(p_{true})}$ & $\frac{\text{mtf\_frq}(\pi_\theta^{dpg})}{\text{mtf\_frq}(r)}$ & $\frac{\text{CE}(T, \pi_\theta^{dis})}{\text{CE}(T, r)}$  & $\frac{\text{mtf\_frq}(\pi_\theta^{dis})}{\text{mtf\_frq}(r)}$  \\ [1mm]  
\hline

\rule{0pt}{3ex}500 & $1.008$ & $1.252$ & $0.76$ & $1.18$ & $281.51$& $0.758$  & $224.94$ \\ [-2mm]
\rule{0pt}{3ex}1000 & $1.014$& $1.102$& $0.762$ &$1.178$ & $240.40$& $0.76$  & $218.24$ \\ [-2mm]
\rule{0pt}{3ex}5000 & $1.019$& $1.21$& $0.865$ &$1.059$ & $34.73$& $0.847$  & $28.69$ \\ [-2mm]
\rule{0pt}{3ex}10000 & $1.014$& $1.067$& $0.968$ &$1.023$ & $2.17$& $0.963$  & $2.04$ \\ [-2mm]
\rule{0pt}{3ex}20000 & $1.004$& $1.023$& $1.0$ &$1.006$ & $1.03$& $1.002$  & $1.01$ \\ [0mm]

\hline\\ [-10mm]
\end{tabular}
\end{center}

\end{table*}

\paragraph{Results}
In Table~\ref{t:stats_dpg_distill_val}
we compute the means of ratios of different quantities across experiments with different motifs, features and seeds: $motif \in \{1000101000101,1011100111001$, $10001011111000\}, ft \in \{ 1001111,Mv1001111\}$, $seed \in \{1234, 4444\}$. In all cases Training-1 is performed using \snis.

These statistics confirm the tendencies illustrated in the previous plots. Namely, when $|D|$ increases the test cross entropy $CE(T,\pit)$ gets closer to the theoretical one $H(\ptrue)$. Also $\pi_\theta$ outperforms $r$ in small conditions of $|D|$ for the two  modes of Training-2: the columns $\frac{\text{CE}(T, \pi_\theta^{dpg})}{\text{CE}(T, r)}$ and $\frac{\text{CE}(T, \pi_\theta^{dis})}{\text{CE}(T, r)}$ show that the models approximate the true process more closely than the initial $r$ in settings with $|D|<10^4$. Similar conclusions can be drawn when comparing the motif frequencies of $\pi_\theta$ and $r$. Further, according to data in columns $\frac{\text{CE}(T, \pi_\theta^{dpg})}{\text{CE}(T, \pi_\theta^{dis})}$ and $\frac{\text{mtf\_frq}(\pi_\theta^{dpg})}{\text{mtf\_frq}(\pi_\theta^{dis})}$, we see that DPG and distillation have comparable efficiency for obtaining the final policy. DPG gives rise to a policy that has better motif frequency but slightly worse cross-entropy than the one from distillation.%\footnote{\TA{what about dis and dpg?}}

% Therefore, distillation and DPG produce AMs that are stronger that original $r$ and as a result our approach is data efficient.\footnote{\TA{this sentence is probably useless}}

%%%%%% RELATED WORK %%%%%%
%\input{related.tex}

\vspace{-2mm}
\section{Conclusion}

Motivated by the GAM formalism for learning sequential models,%%%%
\footnote{The limitation to \emph{sequential} EBMs is not as serious as it seems. Many objects can be decomposed into sequences of actions, and EBMs over such objects could then be handled in similar ways to those proposed here.}
%%%%
we proposed some RL-inspired techniques for obtaining distributional policies approximating the normalized distribution associated with an energy-based model over sequences. We took some first experimental steps, in controlled synthetic conditions, for confirming that these techniques were working.

While the main algorithm (\dpgoff) proposed here for computing distributional policies is generic in the sense that it only requires a potential $P(x)$ and a proposal $q$, the fact that GAMs intrinsically enclose an autoregressive policy $r$ that can be used to initialize such a proposal is an important advantage.
It should also be observed that the division of work in GAMs between Training-1 and Training-2 helps clarifying a distinction that should be made about training sequential EBMs from data. \cite{gams-conll-2019} already observed that training the representation $\Pl$ could be much easier than extracting an autoregressive model from it.\footnote{There are some extreme situations where the $\Pl$ obtained at the end of Training-1 can perfectly represent the true underlying process, but no policy has a chance to approximate $\pl$. This can happen with features associated with complex filters (e.g. of a cryptographic nature) used for generating the data, which can be easily detected as useful during Training-1, but cannot feasibly be projected back onto incremental policies.} 
If we think in the terms of the current paper, we can further observe that while Training-2 has direct connections to RL (exploiting a given reward to obtain a policy), Training-1 has some similarities to \emph{Inverse RL} \cite{Russell:1998:LAU:279943.279964,Ng:2000:AIR:645529.657801}: deriving a reward \emph{from the training data}, here purely inside a max-likelihood approach. Trying to combine the two aspects in one direct algorithm would only blur the true nature of the problem.

The move from the standard optimization view of RL and the sampling (aka distributional) view advocated here is a natural one. Optimization can be seen as an extreme case of sampling with a low temperature, and the approach to distributional policies developped in our Algorithm~\ref{al:DPG} might be a way for developing stable algorithms for standard RL purposes (a related approach is proposed in \cite{Norouzi:2016}).

Our importation of policy gradient from standard RL to the distributional view only scratches the surface, and another promising line of research would be to adapt methods for local credit assignment, such as actor-critic techniques, to the problem of sampling from an energy-based model.

\paragraph{Acknowledgements} Thanks to Tomi Silander and Hady Elsahar for discussions and feedback.

\clearpage
\bibliographystyle{plain} 
%\bibliography{RELATED.bib,AAAI.bib}

\clearpage
\appendix

%%%%%%%%%%%%%%%%%%%%%%%%%%%%%%%%%%%%%%%%%%%%%%%%%%%%%%%%%%%%%%%
%%%%%%%%%%%%%%%%%%%%%%%%%%%%%%%%%%%%%%%%%%%%%%%%%%%%%%%%%%%%%%%
%%%%%%%%%%%%%%%%%%%%%%%%%%%%%%%%%%%%%%%%%%%%%%%%%%%%%%%%%%%%%%%
%%%%%%%%%%%%%%%%%%%%%%%%%%%%%%%%%%%%%%%%%%%%%%%%%%%%%%%%%%%%%%%
%%%%%%%%%%%%%%%%%%%%%%%%%%%%%%%%%%%%%%%%%%%%%%%%%%%%%%%%%%%%%%%
%%%%%%%%%%%%%%%%%%%%%%%%%%%%%%%%%%%%%%%%%%%%%%%%%%%%%%%%%%%%%%%
%%%%%%%%%%%%%%%%%%%%%%%%%%%%%%%%%%%%%%%%%%%%%%%%%%%%%%%%%%%%%%%
\section{Supplementary Material}

\subsection{Related Work}\label{sec:related}

GAMs have been introduced in \cite{gams-conll-2019}. While that paper already proposes the division of training in two stages, it only considers a distillation method, of limited application, for Training-2. It mentions a possible relation with RL as future work, but does not elaborate, while this is a central focus of the present submission.

Global and Energy-Based approaches to neural seq2seq models have been considered in several works. Among those, \cite{andor-etal-2016-globally} consider transition-based neural networks, and contrast local to global normalization of decision sequences, showing how the global approach avoids the \emph{label bias} problem for tasks such as tagging or parsing. Contrarily to us, they focus on inference as maximization, for instance finding the best sequence of tags for a sequence of words. \cite{Belanger:2016:SPE:3045390.3045495} address a similar class of problems (multi-labelling problems such as sequence tagging), employing an energy-based generalization of CRFs, also focussing on inference as optimization. \cite{KimBengio2016}, similar to us, consider probabilistic generative processes defined through an energy-based model. Their focus is on the generation of non-sequential objects, using GAN-type binary discriminators to train the energy representation on the available data. They do not exploit connections to RL.

Reinforcement Learning approaches for seq2seq problems have also been studied in many works. Among those, \cite{DBLP:journals/corr/RanzatoCAZ15} use an hybrid loss function to interpolate between perplexity (aka cross-entropy) training and reward optimization, with the reward being defined by evaluation measures (such as BLEU in machine translation) differing from perplexity. \cite{bahdanau_actor-critic_2016}, still in a RL-as-optimization framework, and with similar objectives, exploit an actor-critic method, where the critic (value function) helps the actor (policy) by reducing variance. \cite{Jaques-2017} and \cite{Norouzi:2016} attempt to combine log-likelihood (aka perplexity) and reward-based training in a more integrated way. In the first paper the rewards are directly defined by \emph{a priori} scores on the quality of the output, which can be computed not only at training time but also at test time. In the second paper, the way in which the rewards are integrated is done by exploiting a probabilistic formulation of rewards close to ours, but used in a different way, in particular without our notion of proposal distribution and with no explicit connection to energy-based modelling. 
In all these cases, the focus is on inference as optimization, not inference as sampling as in the present submission.

Finally, \cite{bellemare_distributional_2017} use a different notion of ``distributional RL" from ours. During policy evaluation, they replace evaluation of the \emph{mean} return from a state by the evaluation of the full \emph{distribution over returns} from that state, and define a Bellman operator for such distribution. Their goal is still to find a policy in the standard (optimization) sense, but with better robustness and stability properties.

\subsection{Rejection Sampling vs. SNIS in Training-1}

Training-1 consists in training the model $P_\lambda$ on $D$. This is done by first training $r$ on $D$ in the standard way (by cross-entropy) and then by training $\lambda$ by SGD with the formula: 
\begin{align}  \label{eq:grad_lamda}
\nabla_\lambda \log p_{\lambda}(x) = \phi(x) - E_{x \sim p_\lambda(\cdot)}\ \phi(x).
\end{align}
The main difficulty then consists in computing an estimate of the model moments $E_{x \sim p_\lambda(\cdot)}\ \phi(x)$. In experiments, \cite{gams-conll-2019} compares two Monte-Carlo approaches \cite{Robert:2005:MCS:1051451} for addressing this problem: (i) \emph{Rejection Sampling} (rs), using $r$ as the proposal distribution and (ii) \emph{Self-Normalized Importance Sampling} (snis) \cite{bengioadaptive2008}, also using $r$ as the proposal.
% (see (SP) for details).\footnote{\MD{We actually should describe these here, no?}}

Rejection sampling is performed as follows. We use $r(x)$ as the proposal, and $P_\lambda(x) = r(x)\ e^{\lambda \cdot \phi(x)}$ as the unnormalized target distribution; for any specific $\lambda$, because our features are binary, we can easily upper-bound the ratio $\frac{P_\lambda(x)}{r(x)} = e^{\lambda \cdot \phi(x)}$ by a number $\beta$; we then sample $x$ from $r$, compute the ratio $\rho(x) = \frac{P_\lambda(x)}{\beta\ r(x)} \leq 1$, and accept $x$ with probability $\rho(x)$. The accepted samples are unbiased samples from $p_\lambda(x)$ and can be used to estimate model moments.

Snis also uses the proposal distribution $r$, but does not require an upper-bound, and is directly oriented towards the computation of expectations. In this case, we sample a number of points $x_1,\ldots,x_N$ from $r$, compute ``importance ratios'' $w(x_i) = \frac{P_\lambda(x_i)}{r(x_i)}$, and estimate $E_{x \sim p_\lambda(\cdot)}\ \phi(x)$ through $\hat{E} = \frac{\sum_i w(x_i) \phi(x_i)}{\sum_i w(x_i)}$. The estimate is biased for a given $N$, but consistent (that is, it converges to the true $E$ for $N \rightarrow \infty$).

In the experiments with DPG in the main text, we only considered cases where Training-1 is done through \snis. This made sense, as both \snis and DPG are motivated by situations in which sampling techniques such as rejection sampling do not work.
\footnote{It is also interesting to note that both \snis and \dpgoff use importance sampling as the underlying technique.}

\begin{figure}[H]
    \begin{center}
    \includegraphics[ width=0.6\linewidth]{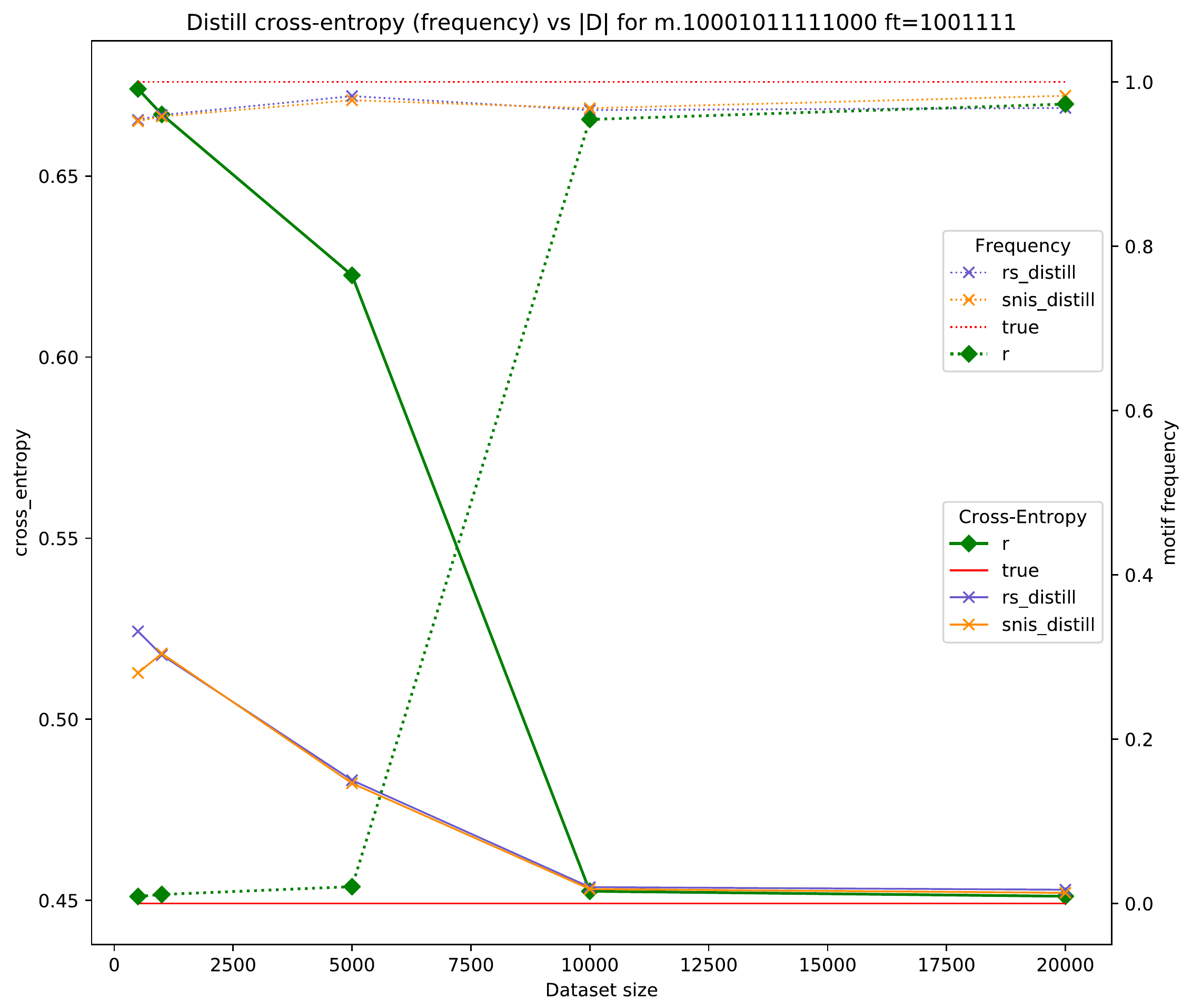} 
    \caption{\label{f:distill_m2s1} \snis vs. \rs for Training-1. In Training-2, only distillation was used.} 
    \end{center}
\end{figure}

Fig.~\ref{f:distill_m2s1} compares \snis with \rs (using only distillation as the Training-2 technique). It can be seen that both techniques produce very similar results. 

\subsection{Effectiveness of DPG in approximating $p$ : details}
\label{supmat:dpg-effectiveness}

\begin{figure}[H]
    \begin{center}
    \includegraphics[width=0.6\linewidth]{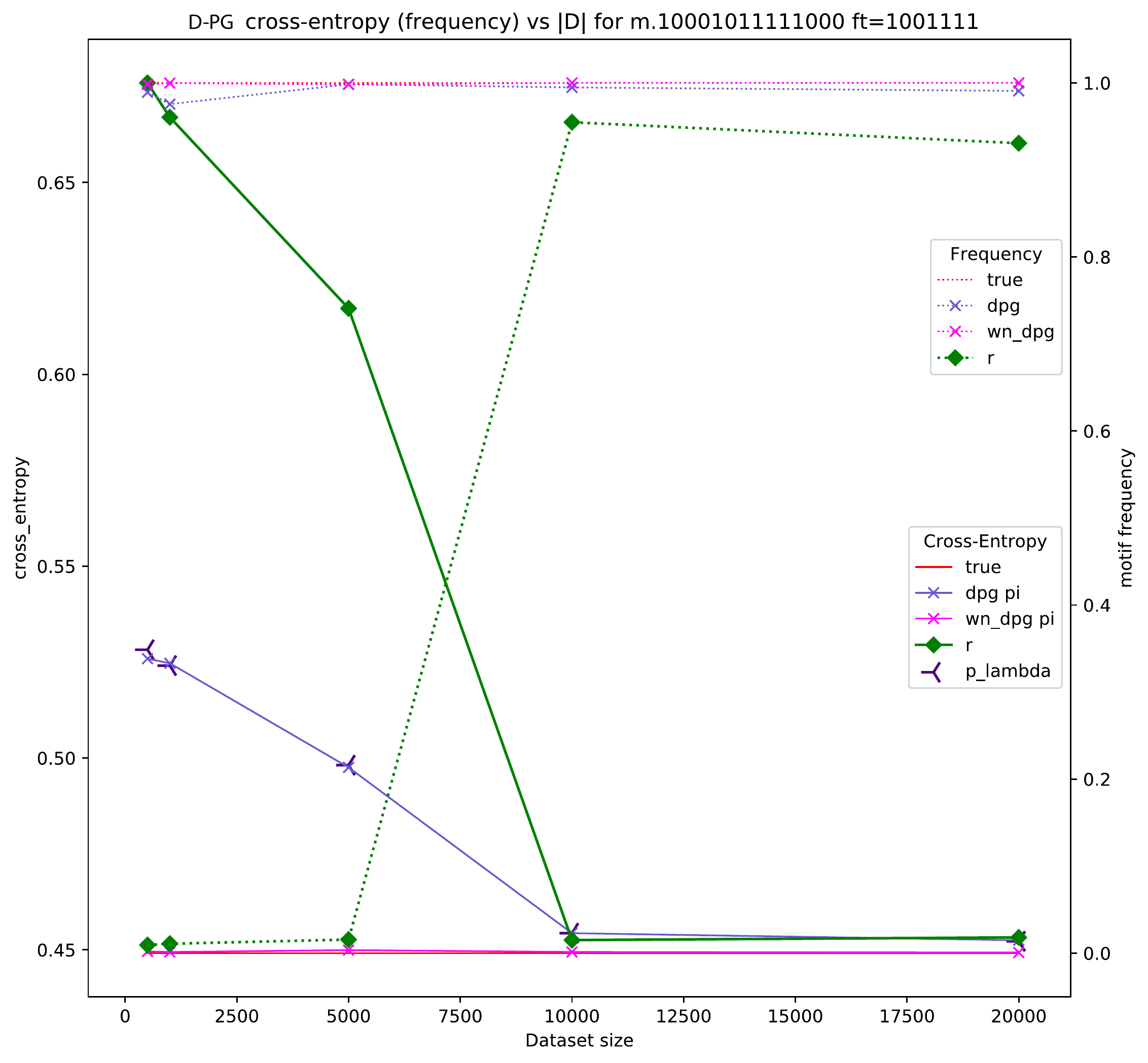} 
    \caption{\label{f:dpg_wndpg} DPG vs. $p$}
    %D-PG vs Wn D-PG. Both $CE(T, \pi_\theta), CE(T, p_\lambda)$ are visualised.}
    \end{center}
\end{figure}

To emphasize the performance of DPG in Training-2 (that is, its effectiveness at finding a distributional policy for an EBM representation $P(x)$), independently of the quality of Training-1), we considered two alternatives for $P$ (see \figurename~\ref{f:dpg_wndpg}). The first one took $P=\Pl$, the energy-based model obtained from Training-1 (the conditions were the same as in \figurename~\ref{f:distill_m2s1}, but we only considered \snis{} for Training-1). For these specific experimental conditions, we were able to accurately estimate (via importance sampling) the partition function $Z$ and therefore to compute the cross entropy $CE(T, p_\lambda)$, represented by the points labelled \texttt{p\_lambda} in the figure. We could then verify that the policy $\pit$ obtained from $\Pl$ by DPG (line \texttt{dpg pl}) was very close to these points. We then considered a second alternative for $P$, namely $P(x)=wn(x) \cdot F(x)$, with $wn(x)$ being the white-noise process filtered by a binary predicate $F(x)$ checking for the presence of the motif; in other words $P(x)$ is an unnormalized version of the true underlying process $\ptrue(x)$. We then applied dpg to this $P$ obtaining the policy represented by the line \texttt{wn\_dpg pl} and we could also verify that this line was very close to the line corresponding to $\ptrue$ (shown as \texttt{true} in the figure, but almost hidden by the other line).

\subsection{Beyond Binary Features in Training-1: Length}
\label{supmat:length-features}

%\footnote{\TA{i think it is important contribution - continuous features! \MD{OK, but not clear where best to place the material, so as not to interrupt message about D-PG}}}

While the emphasis of the current paper is on Training-2 and its relationship with distributional policies in RL, we also wanted to go beyond one of the limiting assumptions of \cite{gams-conll-2019}, namely its reliance on binary features only: e.g., presence of a substring, value of the first bit, etc. We wanted to confirm that GAMs can be applied continuous features as well, and in fact to features that have a strong inter-dependence. We also wanted to consider features that relied on weaker prior knowledge than the presence of specific motifs.

To do that, we considered an additional length feature with two components, namely $\frac{|x|}{\textrm{max\_len}} \in [0,1]$ denoted as $M$ and $\frac{|x|^2}{\textrm{max\_len}^2} \in [0,1]$ denoted as $v$. 

We note that the moments of these two features correspond to sufficient statistics for the normal distribution, and roughly speaking GAMs are obtained by matching moments of the given dataset $D$. 

We were then able during Training-1 to learn the corresponding $\lambda$ parameters using either $snis$ without modification or $rs$ with a modification for computing the upper bound (since the two components are inter-dependent).%\footnote{\TA{the only time when we modify something in training-1 comparing to CONLL}}

\begin{figure}[h!]
    \centering
    \includegraphics[ width=0.65\linewidth]{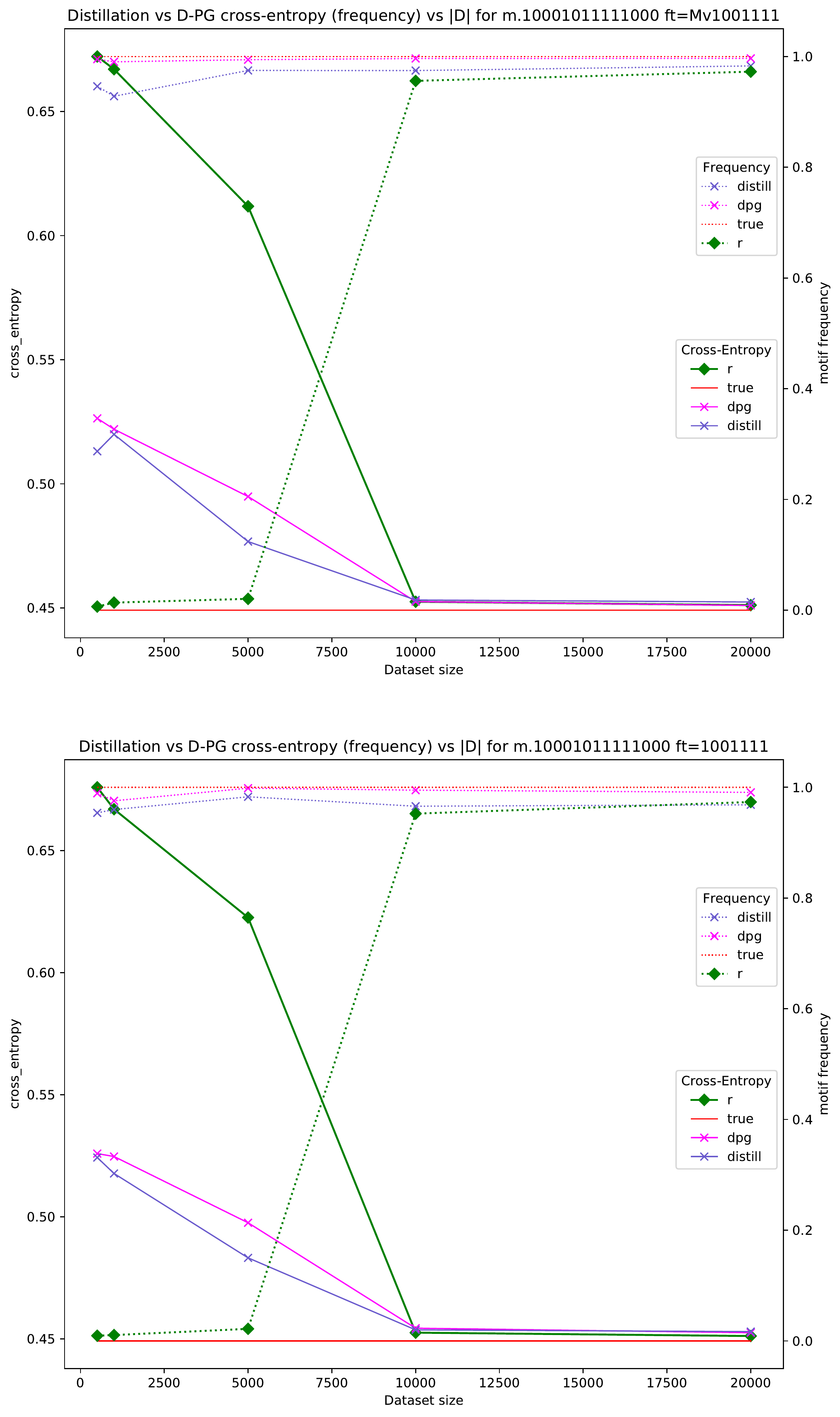} 
    \caption{\label{f:dpg_distill_len} DPG vs Distillation with length feature on (top) %($ft=Mv1001111$) 
    or off (bottom).}
\end{figure}

However, we noticed that the performance of two training setups (distillation and DPG) was rather similar whether the length feature was on or off (see \figurename~\ref{f:dpg_distill_len}). We speculate that in order to see the impact of the length feature, the strings in $D$ should be longer so that the original AM $r$ would be weaker in characterizing the length.

%\clearpage
%\section{WRITING NOTES}
%\theendnotes

\end{document}